\documentclass[conference]{IEEEtran}
\IEEEoverridecommandlockouts
\usepackage{cite}
\usepackage{amsmath,amssymb,amsfonts}

\usepackage{framed,multirow}

\usepackage{latexsym}
\usepackage{booktabs} 
\usepackage{graphicx}
\usepackage{textcomp}
\usepackage{xcolor}
\usepackage{algorithm}
\usepackage{algpseudocode}
\usepackage{subcaption}
\usepackage{comment}
\usepackage{graphicx}
\usepackage{textcomp}
\usepackage{xcolor}
\def\BibTeX{{\rm B\kern-.05em{\sc i\kern-.025em b}\kern-.08em
    T\kern-.1667em\lower.7ex\hbox{E}\kern-.125emX}}
\begin{document}

\title{Prototypical Quadruplet for Few-Shot Class Incremental Learning\\}

\author{\IEEEauthorblockN{Sanchar Palit}
\IEEEauthorblockA{Department of Electrical Engineering \\
Indian Institute of Technology Bombay\\
}
\and
\IEEEauthorblockN{Biplab Banerjee}
\IEEEauthorblockA{Center of Studies in Resources Engineering \\
Indian Institute of Technology Bombay\\
}
\and
\IEEEauthorblockN{Subhasis Chaudhuri}
\IEEEauthorblockA{Department of Electrical Engineering \\
Indian Institute of Technology Bombay\\
}
}

\maketitle

\begin{abstract}
The scarcity of data and incremental learning of new tasks pose two major bottlenecks for many modern computer vision algorithms. The phenomenon of catastrophic forgetting, i.e., the model's inability to classify previously learned data after training with new batches of data, is a major challenge. Conventional methods address catastrophic forgetting while compromising the current session's training. Generative replay-based approaches, such as generative adversarial networks (GANs), have been proposed to mitigate catastrophic forgetting, but training GANs with few samples may lead to instability. To address these challenges, we propose a novel method that improves classification robustness by identifying a better embedding space using an improved contrasting loss. Our approach retains previously acquired knowledge in the embedding space, even when trained with new classes, by updating previous session class prototypes to represent the true class mean, which is crucial for our nearest class mean classification strategy. We demonstrate the effectiveness of our method by showing that the embedding space remains intact after training the model with new classes and outperforms existing state-of-the-art algorithms in terms of accuracy across different sessions.
\end{abstract}


\section{Introduction}

Conventional deep learning models require a large number of labeled datasets to learn meaningful features from the data. These meaningful features are used to do different tasks in Computer vision such as object classification \cite{b4}, object detection \cite{b54},  segmentation \cite{b53} etc. In absence of a proper sufficient dataset, the model might not be able to learn proper representation. If the number of samples available per class is sparse in nature the learner tends to overfit in the standard supervised learning technique. Another drawback of the standard supervised training model is that it needs to learn meaningful features of the data in a single training step. But in a real-life scenario, our full dataset might not be available initially and might arrive sequentially, posing a unique challenge. In such cases, if we do multiple passes over diverse datasets on the learner, it will tend to underperform.

In neural network terminology, these two difficulties have been addressed by several works to learn the representation prudently. Several Few shot class incremental learning (FSCIL) frameworks \cite{b37}, \cite{b56} use \textit{architectural modification} to learn feature space topologies for knowledge representation. The network keeps growing at each incremental session to learn the feature of the new class. This implementation requires one to maintain a separate knowledge base which might be cumbersome in some applications. Several other methods such as \cite{b59}, \cite{b7}, \cite{b57} have used additional \textit{synthesized previous samples} via generative models or additional memory to use the previously learned data. \cite{b8}, \cite{b60}, \cite{b61} relies on \textit{fine-tuning} a part of the base model to mitigate catastrophic forgetting. 

During the training session, \cite{b8}, \cite{b60}, and \cite{b61} have decoupled the feature extractor and the final classification layer which is erroneous because when the feature extractor changes the previous classification weight must change as well. Because along with the feature extractor, all weights at the final classification must be updated to maintain synchronization. Otherwise, the network outputs will change uncontrollably, which is equivalent to forgetting some parameters. In the case of \cite{b59}, \cite{b7}, \cite{b57} additional overhead is in maintaining additional generative modeling (VAE or exemplar samples which are generated using the herding process or maintaining the additional Vector Quantizer model respectively). In addition to that in the case of highly imbalanced classes, the freshly generated sample might not be able to produce true class representatives.


\begin{figure*}[!htb]
\centerline{\includegraphics[width=17cm,height=9cm]{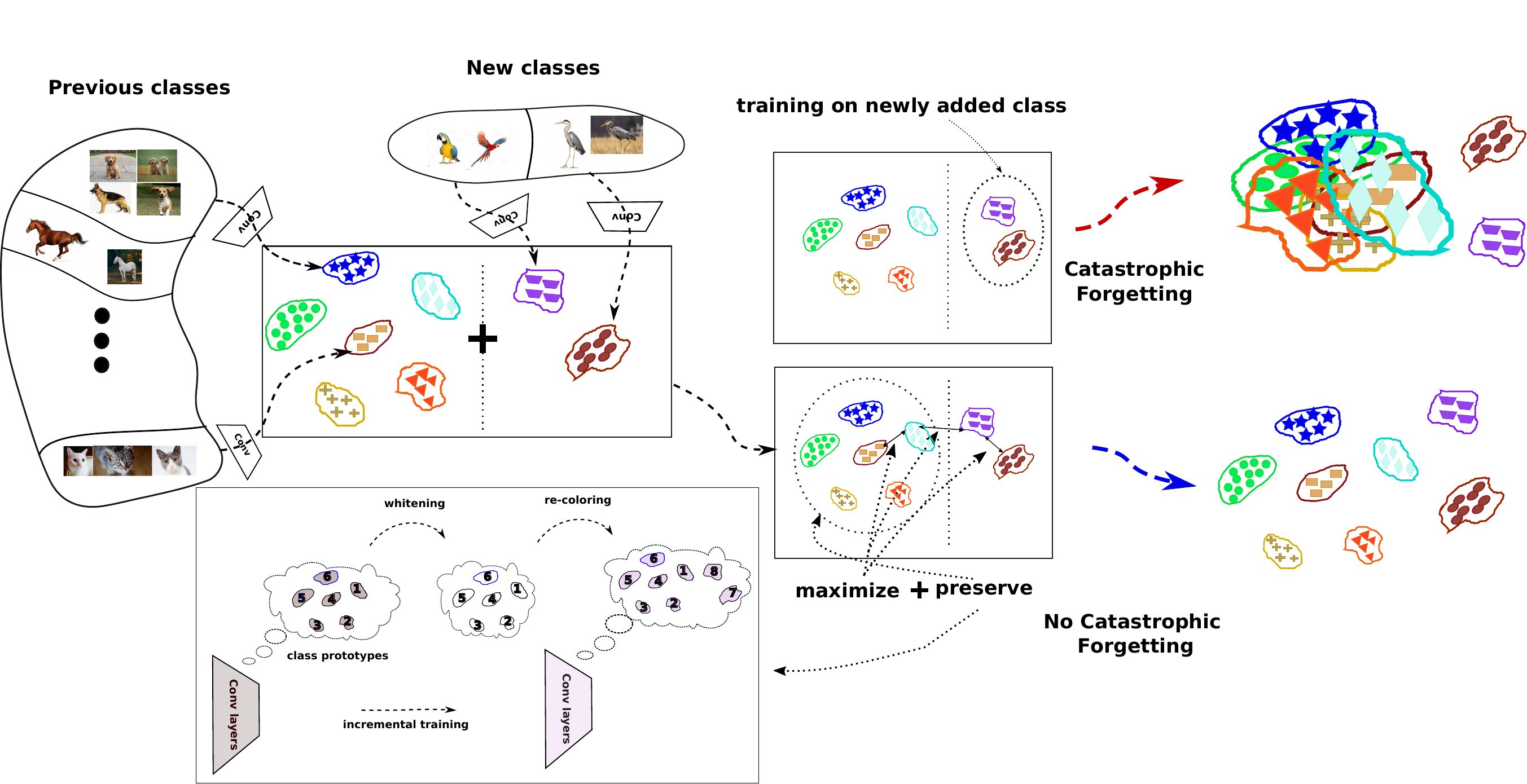}}
\caption{Our contrastive loss learns the current session parameters faithfully while retaining the knowledge of the past. The red upper arrow indicates forgetting while learning new classes the network parameters have changed and the previously learned class clusters have overlapped. In our method (blue lower arrow), the newly learned class cluster is far apart from the previously formed class clusters and the previously formed clusters retain their inter-class distance indicating retaining knowledge of the past.
}
\label{fig:teaser}
\end{figure*}

 In our work, we have tried to focus on preserving knowledge of the past without disregarding the current session parameter learning. We have preserved the previous knowledge with previously seen class prototypes along with assigning enough focus on updating the current session parameter. The previous class prototypes are updated as the parameters are updated during the current training. Our model is able to \textit{distinguish the samples of a class} from,  $(a)$ \textit{previous session class samples} and $(b)$ \textit{samples of other classes of the current session} very efficiently by means of quadruplet loss. We also focus on improved contrastive learning in the current session (see Figure \ref{fig:teaser}). We developed a methodology to get a better embedding in the feature space using the quadruplet loss. The newly sampled data is very similar to the sample mean of the class that it belongs to. Hence we also penalize the distance between the sample and its class mean with a prototypical loss function. In addition to that, we retain the previous knowledge by maintaining the invariability of the previously seen class means. The previously learned class prototypes are refreshed and updated during training at the current session in such a way that those are able to represent the true class means although the network parameter keeps changing from time to time. We keep track of the initial prototype (the prototypes which are created for the first time) and some copies of the given class prototypes. 
 The contribution of the paper is summarized below,
 \begin{itemize} 
 \item We use prototypical quadruplet loss to enlarge the disparity of the different class samples and to improve the similarity of the similar class samples. This creates a \textit{discriminative embedding space}.
 \item We \textit{update the prototypes of each previously seen class} during the training. We use flattened statistics of the mean and variances of the class prototypes via a smoothing kernel. This flattened statistic is used to adjust the prototypes from the previous classifier to the current classifier using the whitening and re-coloring procedure. 
 \item We also ensure that \textit{the prototypes do not deviate far apart from the initial prototype}. In addition to that, we ensure that the \textit{inter-class separability} of the updated prototype is maximized by minimizing the cross-correlation between the updated prototypes. 
\end{itemize}

\section{Related Work}\label{SCM}
 \subsection{Metric learning:}
Metric learning has been used in several few-shot learning-based methods \cite{b3}. Metric-learning-based models define the similarity of data points conditioned on some metric defined on the samples during the training process. An example of a metric is the euclidean distance to the nearest class prototypes in embedding space \cite{b1}, relation modules \cite{b10}, model parameters \cite{b14} or cosine similarity \cite{b11}. Some meta-learning models \cite{b14}, train the parameters of the model recursively over a collection of tasks/samples with a small number of gradient steps with an extra computational cost to produce good generalization performance on that task. Our results show that a simple \textit{similarity-dissimilarity empowered quadruplet method along with a distance-based classifier} (without using any other metric for training over a collection of tasks/episodes) achieves competitive performance with respect to other sophisticated algorithms.

\subsection{Rehearsal and Pattern Replay:}
Several methods \cite{b7}, \cite{saha2021gradient} have stored a fraction of previously learned data and replayed them while learning the current samples. The episodic memory \cite{saha2021gradient}, \cite{lopez2017gradient}, is limited by imposing proper selection and discard criteria. Alternatively, previous training class samples have been generated using GANs \cite{cong2020gan} with additional training costs. Some methods \cite{b13} augment initial training by adding a set of generated examples while training the model end-to-end along with the classification algorithm(a composition of the inference and learning algorithms). Our method avoids rehearsal and pattern replay as it incurs extra overhead while training and we do not use any previous training data. Our method only needs the identifier of each class in the form of a prototype due to which the memory size is very smaller than the compared methods. We reformulate \textit{our knowledge of the past by decorrelating the prototypes from the previous model and correlating them to the current model} (via whitening and recoloring) as the network parameters changes.  

\subsection{Few-Shot Class Incremental Learning(FSCIL):}
  FSCIL have been addressed recently by several works \cite{b8}, \cite{b56}, \cite{b57} for tackling Class Incremental Learning (CIL). Task-incremental learning, which is a generalization of CIL  has been tackled by different settings \cite{b32}. Several attempts have been taken such as a neural Gas network \cite{b37}, semantic word embedding \cite{b58}, and vector quantization \cite{b57}. \cite{b34} have used retrospection along with distillation where retrospection allows the model to revisit a small subset of data for old tasks. \cite{b35} uses the adverse effects caused by the imbalance between old and new classes to tackle catastrophic forgetting. We have used three types of loss in the incremental session viz. \textit{prototypical-quadruplet loss computed on all classes, less-forget constraint, a cosine similarity loss computed on the previous class prototypes with their initial footprint, and inter-class separation, a cross-correlation loss to maintain the inter-class separability of the previous class prototypes}.

\section{Problem Definition}

In FSCIL setting we receive sequentially set of examples $\mathcal{D}^{(1)}, \mathcal{D}^{(2)}, ..,
\mathcal{D}^{(i)}, .., \mathcal{D}^{(S)}$ where $\mathcal{D}^{(i)} = (X_i,y_i)$ and $X_i = \{x_{i,j}\}_{j=1}^{|\mathcal{D}^{(i)}|}$ are input samples and $y_i = \{y_{i,j}\}_{j=1}^{|\mathcal{D}^{(i)}|}$ are output classes such that each $y_{i,j}$ is assigned to the input $x_{i,j} \in \mathbb{R}^D$. The labels $y_i \in \mathcal{N}^{(i)}$ are mutually exclusive i.e for any two steps $\forall k\neq l, \mathcal{N}^{(k)} \cap \mathcal{N}^{(l)} =\phi $. At this point, we should note that at the very first training set $\mathcal{D}^{(1)}$ have more classes and we refer to this step as the ``\textit{base session training}" step. At each subsequent step, the training sets $\mathcal{D}^{(t>1)}$ have an equal number of classes with few samples distributed across all steps which are referred to as ``\textit{incremental session training}" step. Each incremental session dataset contains $n$-number of classes $(n = |\mathcal{N}^{(t)}|)$ and $k$-samples per class. This is termed as ($n$-way, $k$-shot) FSCIL setting where $|\mathcal{D}^{(t)}| = n \times k$. During training at session $t$ only data $\mathcal{D}^{(t)}$ is available and the previous session sets $\mathcal{D}^{(1)}, \mathcal{D}^{(2)}, ... \mathcal{D}^{(t-1)}$ data are not available. After training the model is tested on all previously encountered classes and current session classes $\mathcal{Z}^{(1)}, \mathcal{Z}^{(2)}, ..., \mathcal{Z}^{(t)}$.

\section{Proposed Method: Prototypical Quadruplet}

 \subsection{Architecture:}
A prototypical quadruplet consists of three main parts: a feature extractor module, a fully connected final classification layer, and a prototype memory bank which will be described in section \ref{prototype_update}. We consider a deep convolutional neural network as a feature extractor which is providing the feature vector from an input image: $f_{\theta}(\mathcal{X}):\mathcal{X} \rightarrow \mathbb{R}^M$, where $\mathcal{X}$ is the input image and $\theta$ are the model parameters. This representation is shared across sessions and $\theta$ is a session-shared parameter. To classify a sample to a specific class at $i$-th session we additionally construct an output layer: $\mathcal{O}(f_{\theta}(\mathcal{X}); \mathcal{W}):\mathbb{R}^M \rightarrow \mathbb{R}^d$ with a log softmax activation, where $\mathcal{W} \in \mathbb{R}^d$ are trainable weight vector of linear fully connected layers. We use this layer to classify the non-linear feature vectors obtained by non-linear mapping. This final layer is chosen such that it is able to accommodate all possible classes that can be encountered throughout the training session ($d \approx \bigcup_{i=1}^k {|\mathcal{N}}^{(i)}|$).

\begin{algorithm}
    \caption{\textsc{PrototypeConstruction}: Support set index selection and prototype construction at each episode in an incremental training session. $|\mathcal{D}^{(i)}|$ is the number of examples in the training set and $|\mathcal{N}^{(i)}|$ is the number of classes in the training set. At each episode we select a few $\mathcal{N}^C\leq |\mathcal{N}^{(i)}|$ number of classes. $N_S$ is the number of support examples per class, $N_Q$ is the number of query examples per class. $\textsc{RandomSample}(S,N)$ denotes a set of $N$ elements chosen uniformly at random from set $S$, without replacement.}
    
    \textbf{Input:}Training set $\mathcal{D}^{(i)}$$ =\{(x_{i,1},y_{i,1}),...,(x_{i,|\mathcal{D}^{(i)}|},y_{i,|\mathcal{D}^{(i)}|})\}$, where labels $y_i \in \mathcal{N}^{(i)}$. $\mathcal{D}^{(i,k)}$ denotes the subset of $\mathcal{D}^{(i)}$ containing all elements $(X_i,y_i)$ such that $y_{i,j}=k$ and the prototype set $\mathcal{C}^{(i-1)}$
    
    \textbf{Output:} Positive class and negative class prototype
    \begin{algorithmic} 
        \State $V \gets \textsc{RandomSample}([1,...,|\mathcal{N}^{(i)}|],\mathcal{N}^C)$ 
        \For{$k \in \{1,...,\mathcal{N}^C\}$}
               \State $S_{kp} \gets \textsc{RandomSample}(\mathcal{D}^{(i,V_k)},N_S).$ \hspace*{\fill}
               \State $Q_k \gets \textsc{RandomSample}(\mathcal{D}^{(i,V_k)}\symbol{92} S_{kp},N_Q)$ \hspace*{\fill}
               \linebreak
               \State $k^{\prime} \gets \textsc{RandomSample}( \{1,...,\mathcal{N}^C\}\symbol{92} k,1).$ \hspace*{\fill} 
               \State $k^{\prime\prime} \gets \textsc{RandomSample}( \{1,...,\mathcal{N}^C\}\symbol{92} k\symbol{92}k^{\prime} ,1)$ \hspace*{\fill}
               \linebreak
               \State $\mathbf{c_{kp}} \gets \frac{1}{N_S}\sum_{(x_i,y_i)\in S_{kp}} f_{\theta}(x_i).$ \hspace*{\fill} 
               \linebreak
                \If{$k^{\prime}, k^{\prime\prime} \in \mathcal{N}^{(t)}$} 
                    \State $S_{kn} \gets \textsc{RandomSample}(\mathcal{D}^{(i,V_k^{\prime})},N_S).$ \hspace*{\fill} 
                    \State $S_{knn} \gets \textsc{RandomSample}(\mathcal{D}^{(i,V_k^{\prime\prime})},N_S).$ \hspace*{\fill} 
                   \linebreak
                    \State $\mathbf{c_{kn}} \gets \frac{1}{N_S}\sum_{(x_i,y_i)\in S_{kn}} f_{\theta}(x_i).$
                    \State $\mathbf{c_{knn}} \gets \frac{1}{N_S}\sum_{(x_i,y_i)\in S_{knn}} f_{\theta}(x_i).$ 
                \Else 
                    \If{$k^{\prime}, k^{\prime\prime} \in \mathcal{N}^{(1)}\cup \mathcal{N}^{(2)}\cup .. \cup\mathcal{N}^{(t-1)}$}
                    \State $\mathbf{c_{kn}} \gets \textsc{RandomSample}(\mathcal{C}^{(k-1)},1)$
                    \State $\mathbf{c_{knn}} \gets \textsc{RandomSample}(\mathcal{C}^{(k-1)},1)$
                    \EndIf
                \EndIf
        \EndFor
    \end{algorithmic}
    \label{alg:ALG1}
\end{algorithm}

\begin{algorithm}
\caption{\textsc{IncrementalSessionLoss}:Loss update at each episode in an incremental training session}
\textbf{Input:}Training set $\mathcal{D}^{(i)}=\{(x_{i,1},y_{i,1}),...,(x_{i,|\mathcal{D}^{(i)}|},y_{i,|\mathcal{D}^{(i)}|})\}$ and $\mathcal{C}^{(k)}$.

\textbf{Output:} The loss $\mathcal{L}$ for a randomly generated training episode
\begin{algorithmic}
\State $\mathcal{L} \gets 0$ \Comment{Initialize loss}

\For{$k \in \{1,...,\mathcal{N}^C\}$} 
    \For{$(x, y) \in Q_k$} 
        \State $d_1 \gets d_{euc}(f_{\theta}(\mathcal{X}),\mathbf{c_{kp}}) - d_{euc}(f_{\theta}(\mathcal{X}),\mathbf{c_{kn}}) + \alpha_1 $
        \State $d_2 \gets d_{euc}(f_{\theta}(\mathcal{X}),\mathbf{c_{kp}}) - d_{euc}(\mathbf{c_{knn}},\mathbf{c_{kn}}) + \alpha_2 $
        \State $g \gets d_1 + d_2 $ 
        \State $\mathcal{L} \gets \mathcal{L}+ \frac{1}{\mathcal{N}^CN_Q}\mathcal{L_{PQ}}$ 
        \State $\mathcal{C}^{(k)} \gets \textsc{CalibrateUpdate}(\mathcal{C}^{(k)}, \mu_b, \Sigma_b)$
        \EndFor
\EndFor
\end{algorithmic}
\label{alg:ALG2}
\end{algorithm}

\subsection{Base training:}
 The base training session set $(\mathcal{D}^{(1)})$ contains $\simeq 50\%$ of the classes. This is because we need to train our initial model competitively as the performance of the model in incremental sessions will be highly dependent on the initial model parameters. A Prototypical Quadruplet network is trained on Cross-Entropy loss at the base training step (Algorithm:\ref{alg:ALG_train}). This makes the feature extractor and the final trainable weight vectors robust \cite{9318835} and prepared for the following incremental training. The final classification layer is only trained during the base training session.

\begin{algorithm}
    \caption{Training on sequential data $\mathcal{D}=\{\mathcal{D}^{(1)},···,\mathcal{D}^{(T)}\}$, with $f_{\theta}(\mathcal{X})$ and $\mathcal{W}$}
    
    \textbf{Input:}Training set $\mathcal{D}^{(i)} =\{(x_{i,1},y_{i,1}),...,(x_{i,|\mathcal{D}_i|},y_{i,|\mathcal{D}_i|})\}$, where labels $y_i \in \mathcal{N}^{(i)}$. 
    
    \textbf{Output:} $f_{\theta}(\mathcal{X};\theta)$ and $\mathcal{W}$
    \begin{algorithmic} 
        \For{$i \in \{1,...,T\}$}
        \If{$i = 1$} \Comment{Base session training}
        \State $\mathcal{L} \gets \textsc{CrossEntropy}(\mathcal{D}_i, f_{\theta}(\mathcal{X};\theta), \mathcal{W})$
        \State $(f_{\theta}(\mathcal{X}); \mathcal{W}) \gets \textsc{BackPropagation}(f_{\theta}(\mathcal{X}); \mathcal{W})$
        \EndIf
        \If{$i \geq 1$} \Comment{Incremental session training}
        \State $\mathbf{c_{kp}}, \mathbf{c_{kn}}, \mathbf{c_{knn}} \gets \textsc{ProtoConstruct}(\mathcal{D}_i, \mathcal{C}^{(i-1)})$ 
        \State $\mathcal{L} \gets \textsc{IncrementalSessionLoss}(\mathcal{D}_i, \mathcal{C}^{(k)})$
        \State $f_{\theta}(\mathcal{X};\theta) \gets \textsc{BackPropagation}(f_{\theta}(\mathcal{X};\theta))$
        \EndIf
        \EndFor
    \end{algorithmic}
    \label{alg:ALG_train}
\end{algorithm}

\subsection{Incremental training:}
 At each episode of incremental training session $i$ we are given a small support set \cite{b3} of $|\mathcal{D}^{(i)}|$ labeled  examples $S_k^i = \{(x_{i,1},y_{i,1}),...,(x_{i,|\mathcal{D}^{(i)}|},y_{i,|\mathcal{D}^{(i)}|})\}$ where each $x_{i,j} \in \mathbb{R}^D$ is the $D$-dimensional input vector of an example and $y_{i,j} \in \{1,...,|\mathcal{N}^{(i)}|\}$ is the corresponding label (Algorithm:\ref{alg:ALG1}). $S_k^i$ denotes the set of examples labeled with class $k  \in \{1,...,|\mathcal{N}^{(i)}|\}$ which is specific to session $i$. $|\mathcal{N}^{(1)}|$ denotes the number of classes in the base step and $|\mathcal{N}^{(i>1)}|$ denotes the number of classes in the incremental training session. This support set is then augmented to create a positive support set ($S_{kp}^i$) and two negative support sets ($S_{kn}^i$ and $S_{knn}^i$). The positive class and the query sample class \cite{b3} are the same. The two different negative classes are different from the query class and also from each other. Next query samples ($x_q$) are chosen from the query support set and for each query sample, positive samples ($x_p$) and two different negative samples ($x_m$ and $x_l$) are chosen. These two negative samples are chosen from current session classes as well as from previous session classes. The input quadruplet ($x_q, x_p, x_l$ and $x_m$) are then passed through the non-linear mapping function $f_{\theta}(\mathcal{X})$. If the chosen negative class indices are from the previous session then we don't need to pass them through the feature extractor as we already have the class means. The prototypes are the mean of the embedding feature points belonging to that class.
\begin{equation}
\mathbf{c_k} = \frac{1}{\mid S_k^i \mid } \sum_{(x_{i,j},y_{i,j}) \in  S_k^i} f_{\theta}(x_{i,j})
\end{equation}

\begin{figure*}[!htb]
\centerline{\includegraphics[width=18cm,height=15cm]{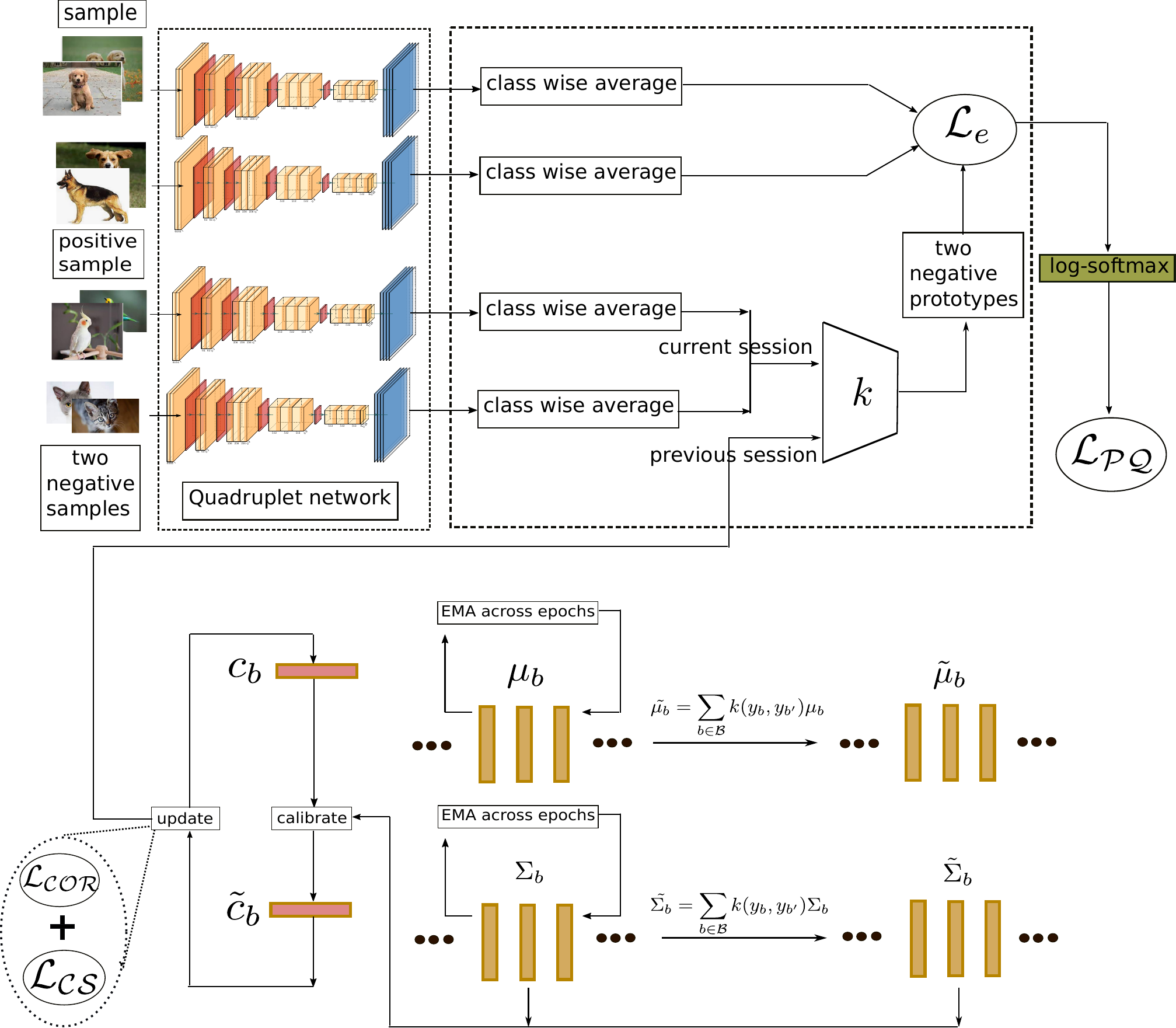}}
\caption{ Prototypical Quadruplet loss update at each incremental training step. At each incremental step, negative prototypes are chosen either from the current session or from the previous session. Previous class prototypes are calibrated and updated during training before using it for prototypical quadruplet loss computation.}
\label{fig:incre}
\end{figure*}

Initially, we create three prototypes i.e. positive class $\mathbf{c_{kp}}$ and two negative class prototypes $\mathbf{c_{kn}}$ and $\mathbf{c_{knn}}$. Given the prototypes $\mathbf{c_{kp}}$, $\mathbf{c_{kn}}$ and $\mathbf{c_{knn}}$ the prototypical quadruplet computes the embedding loss function $\mathcal{L}_e = g_{\mathcal{W}}:\mathbb{R}^M \times\mathbb{R}^M \rightarrow \mathbb{R}^d \times [0,+\infty)$for each query sample as,

\begin{equation}
\begin{split}
\mathcal{L}_e &= g_{\mathcal{W}}(f_{\theta}(x_q),k)=\sum_{x_q\in Query}[d_{euc}(\mathcal{W}^{T}f_{\theta}(x_q),\mathbf{c_{kp}})\\ &- d_{euc}(\mathcal{W}^{T}f_{\theta}(x_q),\mathbf{c_{kn}}) + \alpha_1] + \sum_{x_q\in Query}[d_{euc}(\mathcal{W}^{T}f_{\theta}(x_q)\\ &, \mathbf{c_{kp}}) - d_{euc}(\mathbf{c_{kn}},\mathbf{c_{knn}}) + \alpha_2] 
\end{split}
\label{eq:quad_loss}
\end{equation}

where $ x_q=x_p,x_q\neq x_m,x_q \neq x_l, x_l\neq x_m$,
and $\mathbf{c_{kp}}$ is constructed with samples $x_p$, $\mathbf{c_{kn}}$ is constructed with samples $x_l$ and $\mathbf{c_{knn}}$ is constructed with samples $x_m$. $d_{euc}(a,b)$ denotes the distance between tensors $a,b$ in feature space and $\alpha_1, \alpha_2$ are the loss margins. With the output of the embedding loss function, prototypical quadruplet networks produce a distribution over classes for a query point $x_q$ based on a softmax over distances to the prototypes in the embedding space:

\begin{equation}
 p_{\theta}(y=k \mid x_q) = \frac{exp(-g_{\mathcal{W}}(f_{\theta}(x_q), k))}{\sum_{k^{\prime}} exp(-g_{\mathcal{W}}(f_{\theta}(x_q), k^{\prime}))}
 \end{equation}
 
This is shown in Figure:\ref{fig:incre}. Hence the negative log-likelihood loss, which is used at each episode (Algorithm:\ref{alg:ALG2}), is given as,
\begin{equation}
\begin{split}
\mathcal{L_{PQ}} = -\log p_{\theta}(y=k \mid x_q) = [g_{\mathcal{W}}(f_{\theta}(x_q),k))+ \\
\log\sum_{k^{\prime}}exp(-g_{\mathcal{W}}(f_{\theta}(x_q), k^{\prime}))]
\end{split}
\end{equation}

\subsection{Parameter Selection}\label{AA}

To prevent catastrophic forgetting we use the knowledge of the prototypes of the previously seen classes. The prototypes at the base training session get updated at different steps starting from the base training step. Next in the incremental training session to prevent a drastic change of network parameters $f_{\theta}(\mathcal{X};\theta)$ we should retain those parameters which contributed significantly to the creation of the previously seen class prototypes. To retain the knowledge we freeze \cite{b8} a certain portion of the network parameters which contributed most to the prototype creation i.e. whose values are above a certain threshold (comparison in ablation study). High absolute values of network weight indicate that this portion contributed significantly to the creation of the tensor in the latent space of the previously seen classes. 
At the next session when new mutually exclusive classes are encountered for training, only a fraction of the parameters of the feature extractor is trained on the newly encountered classes. This is because the number of samples in the newly encountered classes is very low, and the entire feature extractor will be prone to overfitting.



\subsection{ Prototype Update}
\label{prototype_update}
In an incremental learning setting, we can not approximate the true class means of the previously seen classes as those would be outdated as the network gets updated at each incremental session. 
In addition to that, we want to improve the inter-class separability of the newly coming class with the previous prototypes. Hence to update the previous means we introduce an additional prototype distribution smoothing \cite{b51} which performs distribution smoothing in the feature space to get the most appropriate class prototypes from the previous prototypes. This procedure aids to adjust the likely skewed approximation of prototype for \textit{underrepresented}, \textit{highly imbalanced} target values in training data. In order to do that we keep track of the $\mathcal{B}$ previously-stored class prototypes. With the prototype statistics, we apply a symmetric kernel $k(\mathbf{c_b},\mathbf{c_{b^\prime}})$ flatten the distribution of the prototype and covariance over the past  $\mathcal{B}$ set of prototypes. Here we should note that we will not have $\mathcal{B}$ copies of all the previous class prototypes. When we are at the $k$-th step, classes $\mathcal{N}^{(1)}, \mathcal{N}^{(2)}, ..,  \mathcal{N}^{(k-\mathcal{B})}$ will have $\mathcal{B}$ copies. Again classes $\mathcal{N}^{(t-\mathcal{B}+1)}$ are smoothened over $(\mathcal{B}-1)$ copies, classes $\mathcal{N}^{(t-\mathcal{B}+2)}$ are smoothened over $(\mathcal{B}-2)$ copies and so on.

To keep track of the deviation of the prototypes we maintain additional mean and covariance matrix $(\mu_b, \Sigma_b)$ which keeps track of the deviation of the prototype of past $\mathcal{B}$ timesteps. The mean and covariance matrix is given as,
$ \mu_b = \frac{1}{\mathcal{B}}\sum_{b\in \mathcal{B}}c_b$ and   $\mathbf{\Sigma_b} = \frac{1}{\mathcal{B}}\sum_{b \in \mathcal{B}}(c_b -\mu_b)(c_b -\mu_b)^{T} $. Next we apply the kernel $k(\mathbf{c_b},\mathbf{c_{b^\prime}})$ on the prototype mean and covariance over the set $\mathcal{B}$. This gives us flattened statistics:
$\Tilde{\mu}_{b} = \sum_{\mathbf{b} \in \mathcal{B}}k(y_b, y_{b^\prime})\mu_b,$ and 
$ \Tilde{\Sigma}_{b} = \sum_{\mathbf{b} \in \mathcal{B}}{k(y_b, y_{b^\prime})\Sigma_b}$. 
Next to adjust the prototype samples we use the standard whitening and re-coloring procedure \cite{b52} with $\{\mu_b, \Sigma_b\}$ and corresponding flattened version $\{\Tilde{\mu}_b, \Tilde{\Sigma}_b\}$:
$ \Tilde{\mathbf{c}}_{b} = \Tilde{\Sigma}_{b}^{\frac{1}{2}}\Tilde{\Sigma}_{b}^{-\frac{1}{2}}(c_b-\mu_b)+ \Tilde{\mu}_b$. Newly constructed ${\mathcal{C}}^{(k)} = ({\Tilde{\mathbf{c}}_1}, {\Tilde{\mathbf{c}}_2}, ..., \Tilde{\mathbf{c}}_{|\mathcal{N}^{k}|})$ is then appended to the previous $\mathcal{B}$ copies of prototypes and the oldest copy is removed. Similarly $\{\Tilde{\mu}_b, \Tilde{\Sigma}_b\}$ is appended to the mean end covariance set and the oldest copy is removed. While training the model across each epoch we implement a momentum update of the running statistics $\{\mu_b, \Sigma_b\}$. Correspondingly, the smoothened statistics $\{\Tilde{\mu}_b, \Tilde{\Sigma}_b\}$ are modified at different epochs but stable within each training epoch. To get a more solid and precise estimate of the prototypes we employ the momentum update which deploys an exponential moving average.

The inter-class separability of the updated prototype is maximized by minimizing the cross-correlation between the updated prototypes:

\begin{equation}
    \mathcal{L_{COR}}(\mathcal{C}^{(k)}) = \sum_{i, j=1, i \neq j}^{|\Tilde{\mathcal{N}}^{k}|}\sigma(\cos{(\tanh{(\Tilde{\mathbf{c}}_{i}^{(k)})}, \tanh{(\Tilde{\mathbf{c}}_{j}^{(k)}}))})
\end{equation}

The other objective is that we need to keep the updated prototypes as close as possible to the previous prototypes. In order to do that we keep track of the initial footprint of the prototypes in a vector $\mathcal{C}^{(0)}$ which stores each class prototype when those are formed initially. To keep the similarity we introduce an additional cosine similarity loss as follows:
\begin{equation}
    \mathcal{L_{CS}}(\mathcal{C}^{(k)}, \mathcal{C}^{(0)}) = \sum_{i=1}^{|\Tilde{\mathcal{N}}^{k}|} \cos{(\tanh{\Tilde{\mathbf{c}}_{i}^{(k)}}, \tanh{\mathbf{c}_{i}^{(0)}})}
\end{equation}
The $\tanh$ is used in addition to the cosine similarity loss because it is sensitive to small deviations. Hence the prototypes are updated at each iteration whereby one update is given as,
\begin{equation}
\mathcal{C}^{(k+1)} = \mathcal{C}^{(k)} -  \lambda\frac{\partial (\mathcal{L_{COR}}(\mathcal{C}^{(k)})+\mathcal{L_{CS}}(\mathcal{C}^{(k)}, \mathcal{C}^{(0)}))}{\partial \mathcal{C}^{(k)}} 
\end{equation}

The prototype update is illustrated in Figure:\ref{fig:incre}.

\subsection{Resource usage}
 At any timestamp during its run time, its memory budget is of the size of the prototype set $\mathcal{B}$ times the number of classes seen so far times the size of the class prototypes. Same number of $\{\mu_b, \Sigma_b\}$ pair is stored. The prototype database i.e. $\Tilde{\mathcal{C}}^{(k)} = \bigcup_{i=1}^{k}{\mathcal{C}}^{(k)}$ consists of $\mathcal{K}=(\mathcal{B}|\mathcal{N}^{(1)}|+ \mathcal{B}|\mathcal{N}^{(2)}|+ ...+ 2|\mathcal{N}^{(|\Tilde{\mathcal{N}}^{k}|-2)}|+|\mathcal{N}^{(|\Tilde{\mathcal{N}}^{k}|-1)}|+  |\mathcal{N}^{(|\Tilde{\mathcal{N}}^{k}|)}|)$ number of $M$-dimensional vectors i.e, $\Tilde{\mathcal{C}}^{(k)} \in \mathbb{R}^{M\times{\mathcal{K}}}$ where $|\Tilde{\mathcal{N}}^{k}|$ denotes the number of classes seen so far i.e $|\Tilde{\mathcal{N}}^{k}|= \bigcup_{i=1}^k {|\mathcal{N}}^{(i)}|$. If we have an estimate of the number of classes that we are going to see next, we can simply allot space for the weight vectors initially and then store $\mathcal{B}$ copies of prototypes in the remaining memory. 

\subsection{Classification} 
The final classification rule after the fully connected layer is $y_{i,j}^{\star} = \underset{\{y_{i, j}\}_{j=1}^{|\mathcal{D}_i|}}{\arg\max}$ $\mathcal{O}(x_{i,j};\mathcal{W}) = \underset{\{y_{i, j}\}_{j=1}^{|\mathcal{D}_i|}}{\arg\max}$ $\mathcal{W}^Tf_{\theta}(x_{i,j})$. The network’s classification method is equivalent to the use of linear fully connected layers with weight vectors $\mathcal{W}$ with non-linear feature vectors obtained by some non-linear mapping. In this process, the non-linear feature map extractor is differentiated from the final linear fully connected layers during training. So we use the nearest class prototype rule \cite{b39} which does not have differentiated weight vectors of the fully connected layers. The class prototypes change in coherence with the non-linear feature map change. This makes the classifier robust whenever there is a change in the non-linear feature map or the feature extractor. As mentioned earlier the model has a set of prototype vectors stored in the memory for each previously encountered class $ \mathbf{c}_1,...,\mathbf{c}_{|\Tilde{\mathcal{N}}^{(k)}|} $. Next, to get a class label during inference, $\mathcal{Y}^{\star}$, for an input sample, $\mathcal{X}$ it gets the distance of the feature vector of the input sample with the prototypes of all encountered classes and specifies the class label with the closest class prototype:

$\mathcal{Y}^{\star}= \underset{i=1,...,{|\Tilde{\mathcal{N}}^{(k)}|}}{\mathbf{\arg\min}} || f_{\theta}(\mathcal{X})-\mathbf{c}_i ||$

\begin{figure*}
\centering
\begin{subfigure}{0.22\textwidth}
  \subfloat[]{\includegraphics[width=4cm,height=3cm]{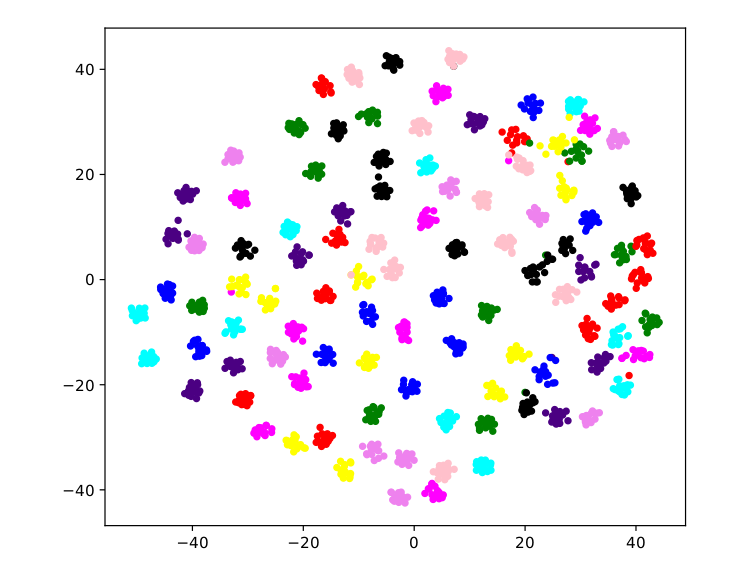}
  \label{fig:sub1_test}}
\end{subfigure}
\begin{subfigure}{0.22\textwidth}
  \subfloat[]{\includegraphics[width=4cm,height=3cm]{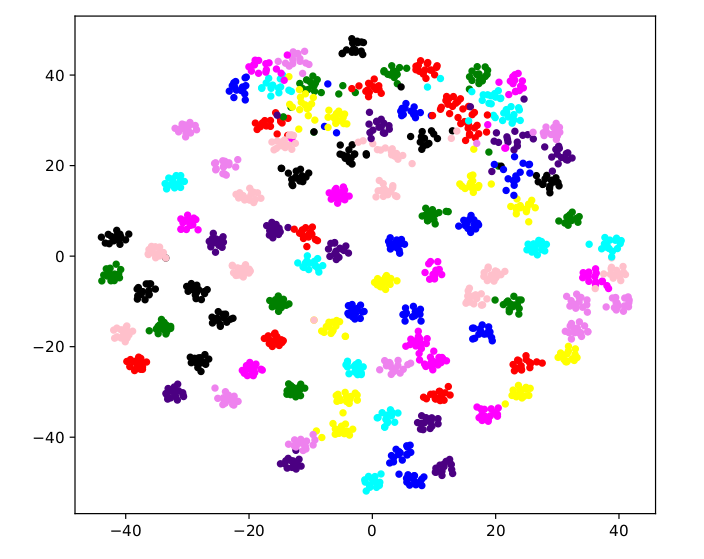}
  \label{fig:sub2_test}}
\end{subfigure}
\begin{subfigure}{0.22\textwidth}
  \subfloat[]{\includegraphics[width=4cm,height=3cm]{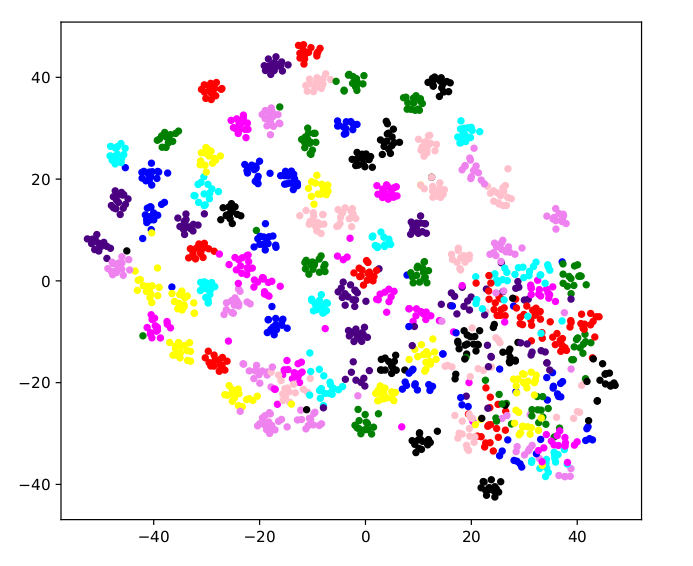}
  \label{fig:sub3_test}}
\end{subfigure}
\begin{subfigure}{0.22\textwidth}
  \subfloat[]{\includegraphics[width=4cm,height=3cm]{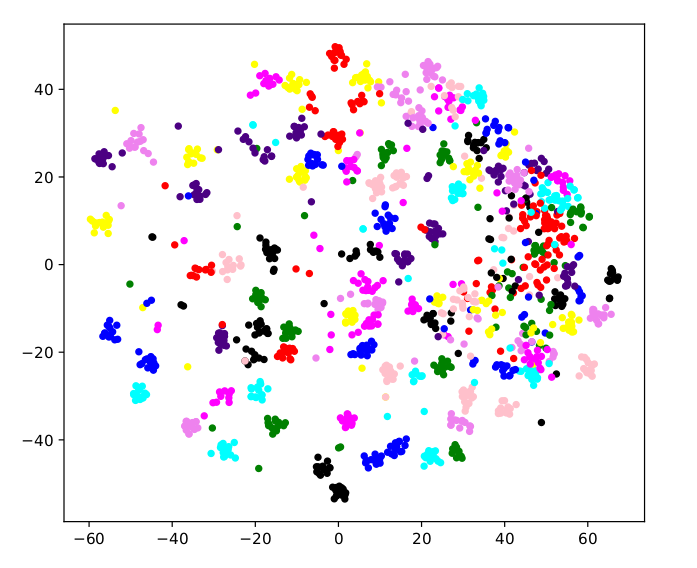}
  \label{fig:sub4_test}}
\end{subfigure}
\caption{t-SNE plot of CUB-200 with prototypical quadruplet network and with (a) $\simeq 10\%$ of the parameters trained, (b) $\simeq 12.5\%$ of the parameters trained, (c) $\simeq 14\%$ of the parameters trained, (d) $\simeq 15\%$ of the parameters trained. All clusters correspond to a different class. Past learned prototypes get deformed due to network weight updates which are crucial in forming the latent space.}
\label{fig:test}
\end{figure*}

\section{Experiments}
We adhere to the standard convention for evaluating the FSCIL method that is suggested in the iCARL\cite{b7}.


\subsection{Datasets}
We use three
image classification datasets CIFAR-100 \cite{b44}, miniImageNet \cite{b11} and CUB200 \cite{b45} to check the performance of our method. 
The dataset partition is shown in Table:\ref{tab_data}
 The test set is comprised of randomly selected test examples that come from all the previously encountered classes.

\begin{table}[htbp]
\caption{Partition of the dataset in base training session and incremental training session:}
\begin{center}
\scalebox{0.7}{
\begin{tabular}{|c|c|c|c|c|c|c|c|c|}
\hline
\textbf{}&\multicolumn{1}{|c|}{\textbf{Base session}}&\multicolumn{7}{|c|}{\textbf{Incremental session}} \\
\cline{2-9} 
\hline
CIFAR-100 & 60 classes & \multicolumn{7}{|c|}{40 classes into 8 sessions with (5-way, 5-shot) setting.} \\
\hline
miniImagenet & 60 classes & \multicolumn{7}{|c|}{40 classes into 8 sessions with (5-way, 5-shot) setting.} \\
\hline
CUB-200 & 100 classes & \multicolumn{7}{|c|}{100 classes into 10 sessions with (10-way, 5-shot) setting.} \\
\hline

\end{tabular}}
\label{tab_data}
\end{center}
\end{table}

\begin{figure*}
\centering
\begin{subfigure}{0.30\textwidth}
  \subfloat[]{\includegraphics[width=6cm,height=6cm]{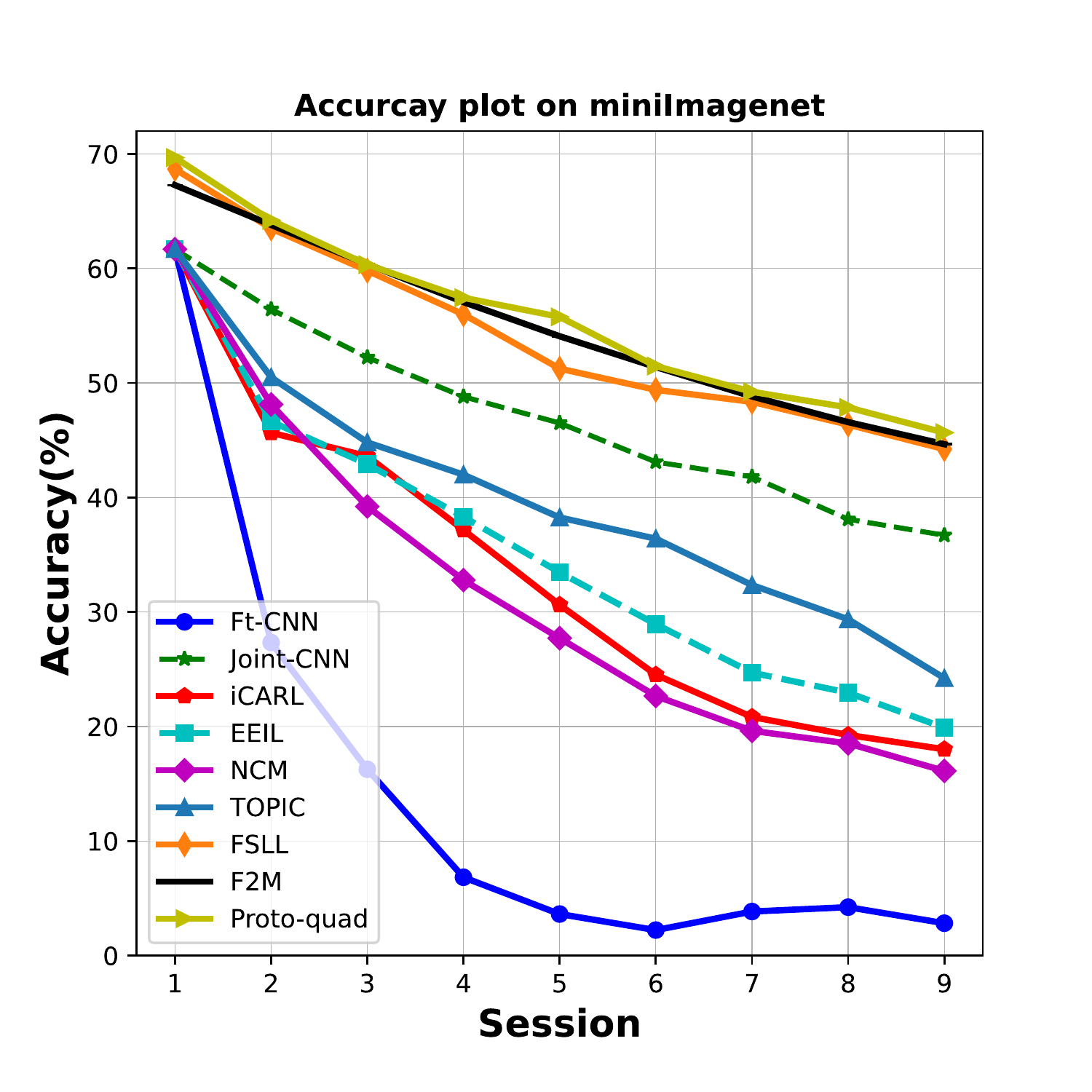}
  \label{fig:miniimage_acc}}
\end{subfigure}
\begin{subfigure}{0.30\textwidth}
  \subfloat[]{\includegraphics[width=6cm,height=6cm]{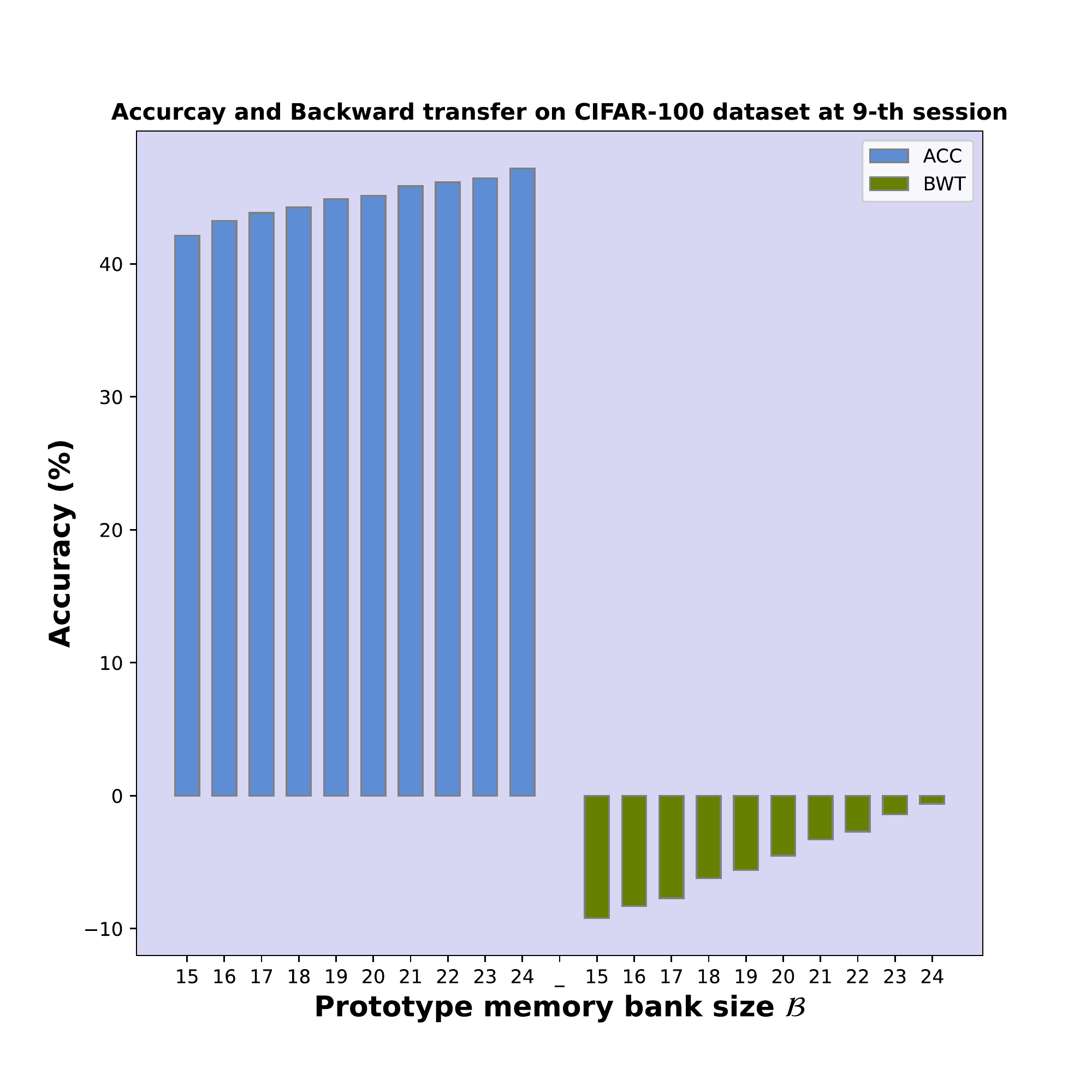}
  \label{fig:proto_bank}}
\end{subfigure}
\begin{subfigure}{0.30\textwidth}
  \subfloat[]{\includegraphics[width=6cm,height=6cm]{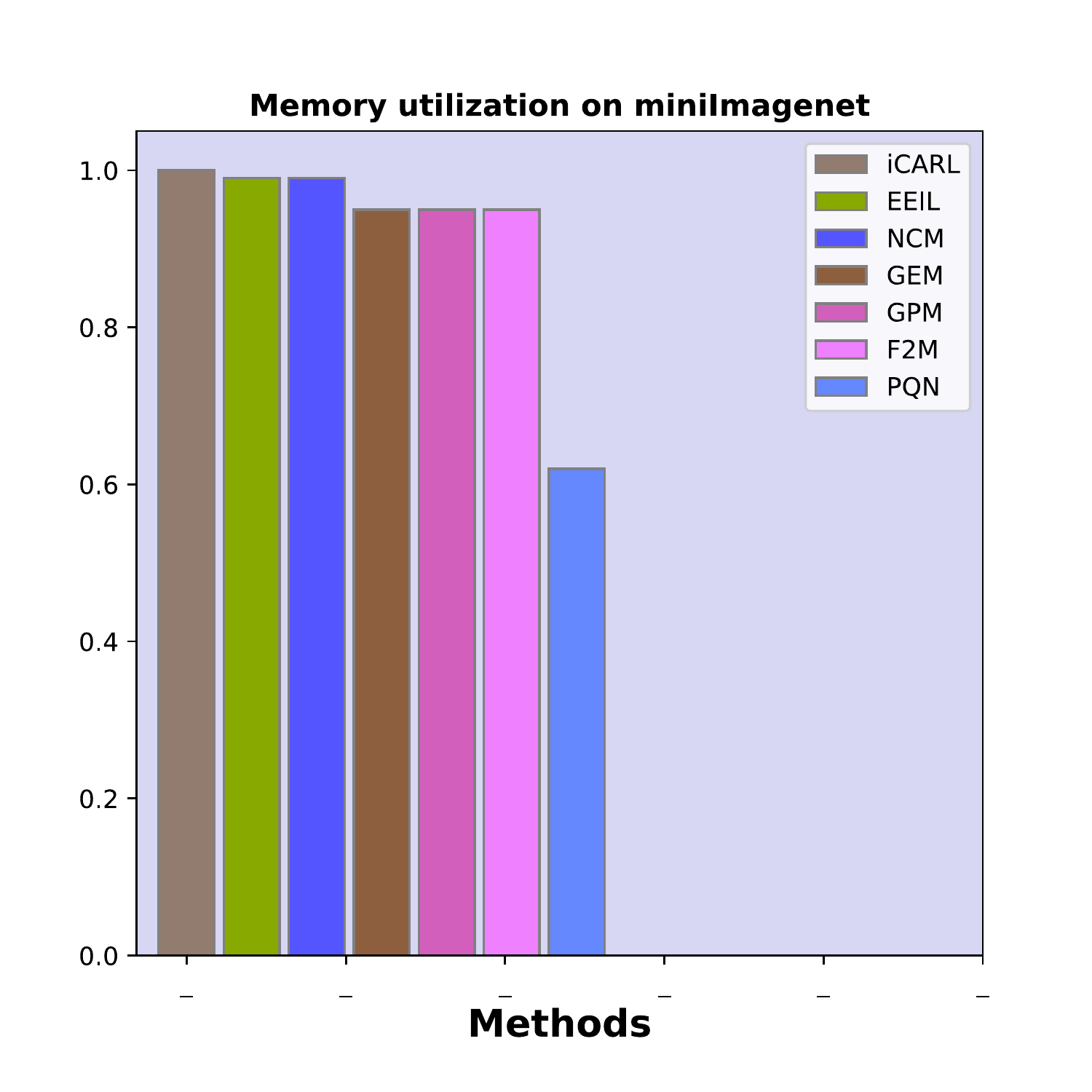}
  \label{fig:mem_util}}
\end{subfigure}
\caption{(a) Results on the miniImagenet on the (5, 5) FSCIL setting. (b) Impact of $\mathcal{B}$ on ACC(\%) and BWT(\%) on CIFAR-100 dataset. With an increasing value of $\mathcal{B}$, the negative BWT reduces, which improves accuracy and reduces forgetting. (c) Memory utilization for different approaches for the miniImagenet dataset. The size of $\texttt{PQN}_{max}$ is
used as a reference (value of 1).}
\label{fig:ablation_mem_acc}
\end{figure*}

\begin{table*}[!htb]
\caption{Results on CUB-200 dataset on the (10, 5) FSCIL setting. At any session, we compute accuracy over all the classes encountered till that session. The final accuracy at the 11th session is reported here.}
\begin{center}
\scalebox{0.85}{
\begin{tabular}{|c|c|c|c|c|c|c|c|c|c|c|c|c|}
\hline
\textbf{Method}&\multicolumn{11}{|c|}{\textbf{Session}} & \textbf{Relative}\\
\cline{2-12} 
\textbf{} & \textbf{1}& \textbf{2}& \textbf{3}& \textbf{4}& \textbf{5}& \textbf{6} & \textbf{7}& \textbf{8}& \textbf{9}& \textbf{10}& \textbf{11}& \textbf{Improvements}\\
\hline
Ft-CNN \cite{b37}& 68.68 & 44.81 & 32.26 & 25.83 & 25.62 & 25.22 & 20.84 & 16.77 & 18.82 & 18.25 & 17.18 & \textbf{46.21}\\
\hline
Joint-CNN \cite{b37}& 68.68 & 62.43 & 57.23 & 52.80 & 49.50 & 46.10 & 42.80 & 40.10 & 38.70 & 37.10 & 35.60 & \textbf{27.79}\\
\hline
iCARL \cite{b7} & 68.68 & 52.65 & 48.61 & 44.16 & 36.62 & 29.52 & 27.83 & 26.26 & 24.01 & 23.89 & 21.16 & \textbf{42.23}\\
\hline
EEIL \cite{b36}&  68.68 & 53.63 & 47.91 & 44.20 & 36.30 & 27.46 & 25.93 & 24.70 & 23.95 & 24.13 & 22.11 & \textbf{41.28}\\
\hline
NCM \cite{b35}& 68.68 & 57.12 & 44.21 & 28.78 & 26.71 & 25.66 & 24.62 & 21.52 & 20.12 & 20.06 & 19.87 & \textbf{43.52}\\
\hline
TOPIC \cite{b37} &  68.68 & 62.49 & 54.81 & 49.99 & 45.25 & 41.40 & 38.35 & 35.36 & 32.22 & 28.31 & 26.28 & \textbf{37.11}\\
\hline

FSLL \cite{b8}& 75.63 & 71.81 & 68.16 & 64.32 & 62.61 & 60.10 & 58.82 & 58.70 & 56.45 & 56.41 & 55.82 & \textbf{7.57}\\
\hline
F2M \cite{hersche2022constrained}& 81.07 & 78.16 & 75.57 & 72.89 & 70.86 & 68.17 & 67.01 & 65.26 & 63.36 & 61.76 & 60.26 & \textbf{3.13} \\
\hline
\textbf{Proto-quad} & \textbf{83.68} & \textbf{80.20} & \textbf{77.33} & \textbf{75.44} & \textbf{74.26} & \textbf{72.47} & \textbf{70.23} & \textbf{68.36} & \textbf{66.28} & \textbf{64.15} & \textbf{63.39} & -\\
\hline
\end{tabular}}
\label{tab1}
\end{center}
\end{table*}

\begin{table*}[!htb]
\caption{Results on CIFAR-100 dataset on the (5, 5) FSCIL setting. At any session, we compute accuracy over all the classes encountered till that session. Final accuracy at the 9th session is reported here.}
\begin{center}
\scalebox{0.8}{
\begin{tabular}{|c|c|c|c|c|c|c|c|c|c|c|}
\hline
\textbf{Method}&\multicolumn{9}{|c|}{\textbf{Session}} & \textbf{Relative}\\
\cline{2-10} 
\textbf{} & \textbf{1}& \textbf{2}& \textbf{3}& \textbf{4}& \textbf{5}& \textbf{6} & \textbf{7}& \textbf{8}& \textbf{9}& \textbf{Improvements} \\
\hline
Ft-CNN \cite{b37}& 64.10 & 38.34 & 15.23 & 10.78 & 8.67 & 5.23 & 5.11 & 3.68 & 2.19 & \textbf{44.98}\\
\hline
Joint-CNN \cite{b37}& 64.10 & 60.15 & 56.42 & 51.43 & 48.87 & 46.10 & 42.80 & 40.10 & 39.70 & \textbf{7.47}\\
\hline
iCARL  \cite{b7} & 64.10 & 53.28 & 41.69 & 34.13 & 27.93 & 25.06 & 20.41 & 15.48 & 13.73 & \textbf{33.44}\\
\hline
EEIL \cite{b36}& 64.10 & 53.11 & 43.71 & 35.15 & 28.96 & 24.98 & 21.01 & 17.26 & 15.85 & \textbf{31.32}\\
\hline
NCM \cite{b35}& 64.10 & 53.05 & 43.96 & 36.97 & 31.61 & 26.73 & 21.23 & 16.78 & 13.54 & \textbf{33.63}\\
\hline
TOPIC \cite{b37}& 64.10 & 55.88 & 47.07 & 45.16 & 40.11 & 36.38 & 33.96 & 31.55 & 29.37 & \textbf{17.8}\\
\hline
FSLL \cite{b8}& 64.10 & 55.85 & 51.71 & 48.59 & 45.34 & 43.25 & 41.52 & 39.81 & 38.16 & \textbf{9.01} \\
\hline
F2M \cite{hersche2022constrained}& 64.71 & 62.05 & 59.01 & 55.58 & 52.55 & 49.96 & 48.08 & 46.67 & 44.67 & \textbf{2.5} \\
\hline
\textbf{Proto-quad} & \textbf{71.10} & \textbf{65.23} & \textbf{60.10} & \textbf{58.34} & \textbf{55.67} & \textbf{52.65} & \textbf{51.19} & \textbf{49.53} & \textbf{47.17} & -\\
\hline
\end{tabular}}
\label{tab2}
\end{center}
\end{table*}

\subsection{Implementation Details}
For CIFAR-100 we use a 32-layers ResNet \cite{b41} feature extractor, and additional memory to store the prototype set of 100 classes. During the base session, the model is trained for 200 epochs and at each incremental session, the model is trained for 60 epochs and 10 episodes. Incremental training is done in an episodic manner\cite{b3}. 
For the CUB-200 dataset, we need to have a prototype set for 200 classes and we use the 18-layer ResNet \cite{b41}. The base training step consists of 200 epochs and each incremental training step consists of 70 epochs. Initially, at each incremental training session, we set the learning rate as 2.0 which is reduced to $1/5$-th after 25, 35, 45, and 55 epochs (5/12, 7/12, 3/4, and 11/12 of all epochs). During the base training session, the model is trained using standard gradient computation and loss update with batches of size 1024, and The weight decay parameter is set to be 0.00001. Here one may notice that the learning rates are large, but they are reasonable with quadruplet loss in the network layer.

\subsection{Baselines and Compared Methods}
For comparative experiments, we run the representative CIL methods in our FSCIL setting, namely iCARL \cite{b7}, EEIL \cite{b36}, NCM \cite{b35}, TOPIC \cite{b37}, F2M \cite{hersche2022constrained} and compare our method with them. Additionally, the Ft-CNN (described in \cite{b37}) and Joint-CNN (also described in \cite{b37} method) are also compared with our method. Ft-CNN directly updates the model on the sparse samples of $\mathcal{D}^{(t>1)}$ and the Joint-CNN updates on the combined samples of $\mathcal{D}^{(1)} \cup \mathcal{D}^{(t>1)}$ i.e base session samples and incremental session samples.

\section{Results and Discussion:}
\subsection{Comparison with other methods:}
 Our method shows better results than the standard model on CUB-200 as indicated by Table:~\ref{tab1}. 
 Moreover, our approach outperforms the F2M method by an amount of 3.13\% in terms of accuracy. Our method shows good accuracy in each incremental training session and the base training session. As shown in Table:~\ref{tab2} our method performs better than the popular F2M method by an amount of 2.5\% on the CIFAR-100 dataset. Similarly, as shown in Figure:\ref{fig:miniimage_acc} our method outperforms the state-of-the-art F2M model, by an amount of 1.01\% on miniImagenet dataset. Table:~\ref{tab2} and Figure:\ref{fig:miniimage_acc} also indicate that our model performs better than other existing methods by a significant amount.

\subsection{Ablation Experiments:}
Several ablation experiments are done to validate our method with different variables.

\subsubsection{Session Trainable Parameters usage}
  The prototypical quadruplet shows the best performance when we update $\simeq 10\%$ parameters of the feature extractor at each incremental session. This is depicted in Figure:\ref{fig:test}. Choosing a more number of session-trainable parameters results in the deformation of the previously learned clusters. For the base step knowledge retention, we decrease the proportion of session trainable parameters and in subsequent steps, the performance of the model improves if we use $\simeq 10\%$ of the feature extractor as session trainable parameters. If we update less than 10\% parameters of the feature extractor, the model starts to show poor performance as it does not have enough parameters to acquire knowledge at the current session.

\subsubsection{Significance of Quadruplet loss margin}
$\alpha_1$ is the margin when we consider the distance between the query and the negative class prototypes along with the query and the positive class prototype. Similarly, $\alpha_2$ is the margin when we consider the distance between the negative class prototype and another negative class prototype.  The threshold margin $\alpha_1$ is imposed between positive and negative samples. It applies an extra border on top of the relative distances between positive and negative class samples w.r.t the same probe negative class sample. The second term enforces new restriction which reviews the orders of positive and negative sample pair with different probe negative class samples. With this restriction, the minimum inter-class separation needs to be more than the maximum intra-class separation notwithstanding of what the pair contains \cite{b43}. We performed ablations to verify the significance of these two margins. We observe that in the absence of these two margins, the session 11 model performance $(S_{11})$ for the CUB-200 dataset drops significantly. Table~\ref{tab4} shows the result with the optimal value of other hyperparameters.

\begin{table}[htbp]
\caption{Session 11($S_{11}$) accuracy results on CUB-200 on the (10, 5) FSCIL setting for different values for the margin hyperparameter $\alpha$ values.}
\begin{center}
\scalebox{0.75}{
\begin{tabular}{|c|c|c|c|c|c|c|c|c|c|}
\hline
\textbf{}&\multicolumn{9}{|c|}{\textbf{Accuracy at session 11}} \\
\cline{2-10} 
\textbf{} & 33.19 & 35.46 & 40.38 & 47.76 & 49.88 & \textbf{57.39} & 52.39 & 51.39 & 50.45 \\
\hline
$\alpha_1$ & 0.5 & 0.75 & 0.7 & 0.8 & 0.9 & \textbf{1.0} & 1.0 & 1.2 & 1.3 \\
\hline
$\alpha_2$ & 1.0 & 0.9 & 0.8 & 0.75 & 0.6 & \textbf{0.5} & 0.3 & 0.2 & 0.1 \\
\hline

\end{tabular}}
\label{tab4}
\end{center}
\end{table}

\subsubsection{Significance of Hyper-parameter $\lambda$}
The update of the prototype plays a significant role along with the session trainable parameters to retain the knowledge. The updated prototype along with the knowledge retention parameters helps retain the knowledge of the previous steps.
We have experimented by changing the prototype update
hyper-parameter $\lambda$ along with the proportion of session trainable parameters on the performance of the model for the CUB-200 dataset in the FSCIL setting. After the $30$ and $40$ epochs, we reduce the learning rate to $\approx10\%$ and $\approx5\%$ of the initial learning rate respectively. We observe the best
model performance for $\lambda =0.1$, and we use this value of the hyper-parameter for all our sessions.  Table~\ref{tab3} shows the result with the optimal value of other hyperparameters.

\begin{table}[htbp]
\caption{Session 11($S_{11}$) classification results on CUB-200 using the ResNet-18 architecture on the 10-way 5-shot FSCIL setting for different values of hyperparameter $\lambda$ values.}
\begin{center}
\scalebox{0.7}{
\begin{tabular}{|c|c|c|c|c|c|c|c|c|c|}
\hline
\textbf{}&\multicolumn{9}{|c|}{\textbf{Accuracy at session 11}} \\
\cline{2-10} 
\hline
$\lambda$ & 0.3 & 0.28 & 0.25 & 0.2 & 0.15 & \textbf{0.1} & 0.08 & 0.07 & 0.05 \\
\hline
Accuracy & 42.44 & 41.06 & 43.42 & 44.27 & 45.33 & \textbf{57.39} & 44.17 & 41.33 & 42.07 \\
\hline

\end{tabular}}
\label{tab3}
\end{center}
\end{table}

\subsubsection{Significance of prototype memory bank size}
Figure:\ref{fig:proto_bank} depicts the effect of increasing the size of the prototype memory bank. For the CIFAR-100 dataset, we have used a variable-size memory bank for each session. We have stored more prototypes for initial sessions (0, 1, 2) and less number of copies of the most recent sessions (6, 7, 8). As a direct consequence, the negative backward transfer (BWT) \cite{saha2021gradient} almost goes down to zero with the increasing size.

While we store these prototypes in memory, we performed a comparative study of memory utilization. Figure:\ref{fig:mem_util} compares the memory utilization of all the memory-based approaches. While \cite{b7}, \cite{b36}, \cite{b35}, \cite{lopez2017gradient}, \cite{saha2021gradient}, \cite{hersche2022constrained} use nearly a memory size of $\texttt{PQN}_{max}$, we obtain a better accuracy by only using 62\% of the $\texttt{PQN}_{max}$.

\section{Conclusion}
We have proposed a novel approach to overcome catastrophic forgetting in few-shot class incremental learning setting by addressing the problem of overfitting with the help of a prototypical network along with quadruplet loss which helps in forming a better embedding space. Extensive experiments on benchmark datasets show that our model can effectively mitigate catastrophic forgetting and adapt to new classes. In our method, one needs to intelligently decide $\mathcal{B}$ set to get robust learning. For future work, one can update the stored prototypes in such a way that those become relevant in the current session and tune the model accordingly during incremental learning sessions, where previous techniques such as elastic weight consolidation (EWC) \cite{b46} can be used to constraint the model parameters.

\bibliographystyle{IEEEtran}
\bibliography{reference.bib}

\end{document}


\title{Supplementary Material}

\maketitle

\section{Algorithm}
Base and incremental training is described in algorithm~\ref{alg:ALG_train}. Prototype construction is described in algorithm~\ref{alg:ALG1} and Incremental session loss computation is described in algorithm~\ref{alg:ALG2}

\begin{algorithm*}
    \caption{\textsc{PrototypeConstruction}:Support set index selection and prototype construction at each incremental training session. $N$ is the number of examples in the training set, $K$ is the number of classes in the training set i.e $|\mathcal{N}^{(i)}|$, $N_C\leq K$ is the number of classes per episode, $N_S$ is the number of support examples per class, $N_Q$ is the number of query examples per class. $\textsc{RandomSample}(S,N)$ denotes a set of $N$ elements chosen uniformly at random from set $S$, without replacement.}
    
    \textbf{Input:}Training set $\mathcal{D}_i =\{(x_{i,1},y_{i,1}),...,(x_{i,|\mathcal{D}_i|},y_{i,|\mathcal{D}_i|})\}$, where labels $y_i \in \mathcal{N}^{(i)} =\{1,..., K\}$. $\mathcal{D}_{i,k}$ denotes the subset of $\mathcal{D}_i$ containing all elements $(X_i,y_i)$ such that $y_{i,j}=k$ and the prototype set $\mathcal{C}^{(i-1)}$
    
    \textbf{Output:} Positive class and negative class prototype
    \begin{algorithmic} 
        \State $V \gets \textsc{RandomSample}([1,...,K],N_C)$ \Comment{Select class indices}
        \For{$k \in \{1,...,N_C\}$}
               \State $S_{kp} \gets \textsc{RandomSample}(\mathcal{D}_{i,V_k},N_S).$ \hspace*{\fill}\Comment{\texttt{positive} support set.} 
               \State $Q_k \gets \textsc{RandomSample}(\mathcal{D}_{i,V_k}\symbol{92} S_{kp},N_Q)$ \hspace*{\fill}\Comment{query set.} 
               \linebreak
               \State $k^{\prime} \gets \textsc{RandomSample}( \{1,...,N_C\}\symbol{92} k,1).$ \hspace*{\fill} \Comment{select indices of \texttt{negative} prototype}
               \State $k^{\prime\prime} \gets \textsc{RandomSample}( \{1,...,N_C\}\symbol{92} k\symbol{92}k^{\prime} ,1)$ \hspace*{\fill}
               \Comment{select indices of another \texttt{negative} prototype}
               \linebreak
               \State $\mathbf{c_{kp}} \gets \frac{1}{N_S}\sum_{(x_i,y_i)\in S_{kp}} f_{\phi}(x_i).$  \Comment{\texttt{positive} prototype}
               \linebreak
                \If{$k^{\prime}, k^{\prime\prime} \in \mathcal{N}^{(t)}$} \Comment{\texttt{negative} prototype from current session}
                    \State $S_{kn} \gets \textsc{RandomSample}(\mathcal{D}_{i,V_k^{\prime}},N_S).$ \hspace*{\fill} 
                    \State $S_{knn} \gets \textsc{RandomSample}(\mathcal{D}_{i,V_k^{\prime\prime}},N_S).$ \hspace*{\fill} 
                   \linebreak
                    \State $\mathbf{c_{kn}} \gets \frac{1}{N_S}\sum_{(x_i,y_i)\in S_{kn}} f_{\phi}(x_i).$
                    \State $\mathbf{c_{knn}} \gets \frac{1}{N_S}\sum_{(x_i,y_i)\in S_{knn}} f_{\phi}(x_i).$ 
                \Else \Comment{\texttt{negative} prototype from past session}
                    \If{$k^{\prime}, k^{\prime\prime} \in \mathcal{N}^{(1)}\cup \mathcal{N}^{(2)}\cup .. \cup\mathcal{N}^{(t-1)}$}
                    \State $\mathbf{c_{kn}} \gets \textsc{RandomSample}(\mathcal{C}^{(k-1)},1)$
                    \State $\mathbf{c_{knn}} \gets \textsc{RandomSample}(\mathcal{C}^{(k-1)},1)$
                    \EndIf
                \EndIf
        \EndFor
    \end{algorithmic}
    \label{alg:ALG1}
\end{algorithm*}

\begin{algorithm*}
    \caption{Training on sequential data $\mathcal{D}=\{\mathcal{D}_1,···,\mathcal{D}_T\}$, with $f_{\phi}(\mathcal{X};\theta)$ and $\mathcal{W}$}
    
    \textbf{Input:}Training set $\mathcal{D}_i =\{(x_{i,1},y_{i,1}),...,(x_{i,|\mathcal{D}_i|},y_{i,|\mathcal{D}_i|})\}$, where labels $y_i \in \mathcal{N}^{(i)} =\{1,..., K\}$. 
    
    \textbf{Output:} $f_{\phi}(\mathcal{X};\theta)$ and $\mathcal{W}$
    \begin{algorithmic} 
        \For{$i \in \{1,...,T\}$}
        \If{$i = 1$} \Comment{Base session training}
        \State $\mathcal{L} \gets \textsc{CrossEntropy}(\mathcal{D}_i, f_{\phi}(\mathcal{X};\theta), \mathcal{W})$
        \State $(f_{\phi}(\mathcal{X};\theta), \mathcal{W}) \gets \textsc{BackPropagation}(f_{\phi}(\mathcal{X};\theta), \mathcal{W})$ 
        \EndIf
        \If{$i \geq 1$} \Comment{Incremental session training}
        \State $\mathbf{c_{kp}}, \mathbf{c_{kn}}, \mathbf{c_{knn}} \gets \textsc{PrototypeConstruction}(\mathcal{D}_i, \mathcal{C}^{(i-1)})$ 
        \State $\mathcal{L} \gets \textsc{IncrementalSessionLoss}(\mathcal{D}_i, \mathcal{C}^{(k)})$
        \State $f_{\phi}(x;\theta) \gets \textsc{BackPropagation}(f_{\phi}(\mathcal{X};\theta))$
        \EndIf
        \EndFor
    \end{algorithmic}
    \label{alg:ALG_train}
\end{algorithm*}

\begin{algorithm*}
\caption{\textsc{IncrementalSessionLoss}:Loss update at each incremental training session}
\textbf{Input:}Training set $\mathcal{D}_i =\{(x_{i,1},y_{i,1}),...,(x_{i,|\mathcal{D}_i|},y_{i,|\mathcal{D}_i|})\}$ and $\mathcal{C}^{(k)}$, where labels $y_i \in \mathcal{N}^{(i)}$. $N_C\leq K$ is the number of classes per episode.

\textbf{Output:} The loss $\mathcal{L}$ for a randomly generated training episode
\begin{algorithmic}
\State $\mathcal{L} \gets 0$ \Comment{Initialize loss}

\For{$k \in \{1,...,N_C\}$} 
    \For{$(x, y) \in Q_k$} 
        \State $d_1 \gets d_{euc}(f_{\phi}(x),\mathbf{c_{kp}}) - d_{euc}(f_{\phi}(x),\mathbf{c_{kn}}) + \alpha_1 $
        \State $d_2 \gets d_{euc}(f_{\phi}(x),\mathbf{c_{kp}}) - d_{euc}(\mathbf{c_{knn}},\mathbf{c_{kn}}) + \alpha_2 $
        \State $g \gets d_1 + d_2 $ 
        \State $\mathcal{L} \gets \mathcal{L}+ \frac{1}{N_CN_Q}\mathcal{L_{PQ}}$ 
        \State $\mathcal{C}^{(k)} \gets \textsc{CalibrateAndUpdate}(\mathcal{C}^{(k)}, \mu_b, \Sigma_b)$
        \EndFor
\EndFor
\end{algorithmic}
\label{alg:ALG2}
\end{algorithm*}

\section{Experiments:}
For the given FSCIL problem, initially, the classes are permuted randomly and then in two parts for the base training set and incremental training respectively. Then incremental training sessions those then divided into different sets. Next at each session, the model is trained distinctly with the training data that is available at the current session. After training the model on the currently available dataset, the trained model then is checked on the test data of the current session set, and also on the test data of all those sets that have already been encountered. Here one can notice that the test data of the previous session is used repeatedly and those are sparse in nature. Still, this does not harm our model since the update on the test data does not change the actual model. The classification accuracies consist of the accuracy on the current session dataset and also on the previous session dataset. Finally, we compute the average of these accuracies which is the final few-shot class incremental accuracy.

\subsection{Dataset:}
The CIFAR-100 dataset consists of
100 classes with each class containing 500 training images
and 100 testing images. Each of the 60,000 images is of
size 32 × 32. The  miniImageNet  dataset  also consists of 60,000 images from 100 classes, chosen from the ImageNet-1k dataset. There are 500 training and 100 test images of size 84×84 for each class. The CUB-200
dataset consists of about 6,000 training images and 6,000
test images for 200 categories of birds. The images are resized to 256×256 and then cropped to 224×224 for training.

\subsection{Implementation details:}
 The features are extracted by a backbone network $f_{\phi}$. Initially $f_{\phi}^{(1)}$ and $\mathcal{W}$ are trained on the base training set $\mathcal{D}_{train}^{(1)}$. The model is trained on $\mathcal{D}_{train}^{(1)}$ for a total of 200 epochs at the base training set. Next, only a small portion of the feature extractor is updated at each incremental training session on each of the sparse training samples $\mathcal{D}_{train}^{(t>1)}$ for 60 epochs. The threshold values for each layer are set to only select $\simeq 10\%$ of $f_{\phi}$ as the parameters to be updated of the feature extractor for each incremental training session. Since the few-shot training sets are very sparse in nature, each episode has only 5 examples from the 10 classes. The feature extractor $f_{\phi}^{t}$ is tested on the cumulative test samples of all the previously encountered classes after finetuning it on the data $\mathcal{D}_{test}^{(t)}$. At any session $t$ we compute accuracy over the test set of current session classes and the classes that are seen at all previous sessions i.e. $(\mathcal{D}_{test}^{(1)}, \mathcal{D}_{test}^{(2)}, ... \mathcal{D}_{test}^{(t)})$ which refers to the session $t$ accuracy. As the number of samples is very few, standard random scaling and rotating for data augmentation \cite{b41}, \cite{b35} is performed for all incremental sessions. In addition, while updating the feature extractor parameters on $\mathcal{D}^{(t>1)}$ the batch-normalization of each layer is done using the batch-normalization statistics computed on base training $\mathcal{D}^{(1)}$ as done in \cite{b37}. We have run these experiments 10 times and the result consists of the average test accuracy over all previous session classes to validate the adaptability and it is seen that standard deviations of the test results are less than 0.5\% for all the experiments.




\section{Ablation Experiments:}
\subsection{Significance of Hyper-parameter:}
The update of the prototype plays a big role along with the session trainable parameters to retain the knowledge. The updated prototype along with the knowledge retention parameters helps retain the knowledge of the previous steps. When the proportion of session trainable parameters is low and the corresponding prototype update should not change the initial values much to retain the values of the previous prototype. 
We have experimented by changing the prototype update
hyper-parameter $\lambda$ along with the proportion of session trainable parameters on the performance of the model for the CUB-200 dataset in the FSCIL setting. After the $30$ and $40$ epochs, we reduce the learning rate to $\approx10\%$ and $\approx5\%$ of the initial learning rate respectively. We observe the best
model performance for $\lambda =0.1$, and we use this value of the hyper-parameter for all our sessions.  Table~\ref{tab3} shows the result with the optimal value of other hyperparameters.

\begin{table}[htbp]
\caption{Session 11($S_{11}$) classification results on CUB-200 using the ResNet-18 architecture on the 10-way 5-shot FSCIL setting for different values of hyperparameter $\lambda$ values.}
\begin{center}
\scalebox{0.7}{
\begin{tabular}{|c|c|c|c|c|c|c|c|c|c|}
\hline
\textbf{}&\multicolumn{9}{|c|}{\textbf{Accuracy at session 11}} \\
\cline{2-10} 
\hline
$\lambda$ & 0.3 & 0.28 & 0.25 & 0.2 & 0.15 & \textbf{0.1} & 0.08 & 0.07 & 0.05 \\
\hline
Accuracy & 42.44 & 41.06 & 43.42 & 44.27 & 45.33 & \textbf{57.39} & 44.17 & 41.33 & 42.07 \\
\hline

\end{tabular}}
\label{tab3}
\end{center}
\end{table}

\subsection{Choice of the loss function}
Our embedding loss function is given as,
\begin{equation}
\begin{split}
g_{\mathcal{W}}(f_\phi(x_q), k)=\sum_{x_q\in Query}[d_{euc}(\mathcal{W}^{T}f_\phi(x_q),\mathbf{c_{kp}})- d_{euc}(\mathcal{W}^{T}f_\phi(x_q)\\,\mathbf{c_{kn}}) + \alpha_1] +  \sum_{x_q\in Query}[d_{euc}(\mathcal{W}^{T}f_\phi(x_q),\mathbf{c_{kp}}) - d(\mathbf{c_{kn}},\mathbf{c_{knn}}) + \alpha_2]
\end{split}
\label{eq:quad_loss}
\end{equation}
We have used quadruplet loss to have a better feature representation in the embedding space because of its relative improvement over the contrastive loss and the triplet loss. Contrastive loss such as Siamese twins pair \cite{chopra2005learning} tries to balance between the loss which is composed of a partial loss function for a genuine pair and the partial loss function for an imposter pair. The triplet loss purely considers the error of the relative distance between positive and negative pairs as long as it exists. But the first term of the contrastive loss gives priority to the absolute distance of positive pairs when the error of the relative distance is not large enough. It would cause the contrastive loss to obtain a small positive distance with the risk of existing errors in the relative distances between positive and negative pairs. Triplet loss is part of quadruplet loss but without the second term in equation~\ref{eq:quad_loss}. The second term provides help from the perspective of orders with different probe images. It can further enlarge the inter-class variations and improve the performance of the testing data. As a result, we can find that quadruplet loss covers the weaknesses of both the binary classification loss and the triplet loss to some extent,  and takes advantage of few-shot class incremental learning which achieves a better performance than either of them.

We also perform experiments with quadruplet loss and contrastive loss as well in the incremental training session for the few-shot training sets $(\mathcal{D}^{(t>1)})$. Our experiments on the CUB-200 dataset show that this results in a session 11 $(S_{11})$ accuracy of 42.13\% for contrastive loss and 44.34\% for triplet loss, which is lower than 57.39\% achieved by quadruplet loss. Therefore, using the quadruplet loss for training on $\mathcal{D}^{(t>1)}$ produces more benefits compared to triplet loss and contrastive loss.

\bibliographystyle{IEEEtran}
\bibliography{supplementary_reference.bib}